\documentclass{article}

\usepackage{arxiv}

\usepackage[utf8]{inputenc} 
\usepackage[T1]{fontenc}    
\usepackage{hyperref}       
\usepackage{url}            
\usepackage{booktabs}       
\usepackage{amsfonts}       
\usepackage{nicefrac}       
\usepackage{microtype}      
\usepackage{lipsum}		
\usepackage{graphicx}
\usepackage{natbib}
\usepackage{doi}
\usepackage{xcolor}
\usepackage{natbib}
\usepackage{array}
\usepackage{makecell}

\usepackage[utf8]{inputenc} 
\usepackage[T1]{fontenc}    
\usepackage{hyperref}       
\usepackage{url}            
\usepackage{booktabs}       
\usepackage{amsfonts}       
\usepackage{nicefrac}       
\usepackage{microtype}      
\usepackage{xcolor}         

\usepackage{graphicx}

\title{DaG LLM ver 1.0: Pioneering Instruction-Tuned Language Modeling for Korean NLP}

\author{
  \href{https://orcid.org/0000-0000-0000-0000}{\includegraphics[scale=0.06]{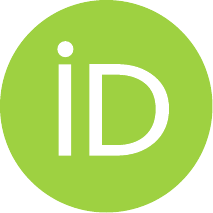}\hspace{1mm}Dongjun Jang} \\
  Department of Linguistics \\
  Seoul National University \\
  \texttt{qwer4107@snu.ac.kr} \\
  \And
  Sangah Lee \\
  Department of Linguistics \\
  Seoul National University \\
  \texttt{sanalee@snu.ac.kr} \\
  \And
  Sungjoo Byun \\
  Department of Linguistics \\
  Seoul National University \\
  \texttt{byunsj@snu.ac.kr} \\
  \And
  Jinwoong Kim \\
  Graduate School of Data Science \\
  Seoul National University \\
  \texttt{kjw900106@snu.ac.kr} \\
  \And
  Jean Seo \\
  Department of Linguistics \\
  Seoul National University \\
  \texttt{seemdog@snu.ac.kr} \\
  \And
  Minseok Kim \\
  Department of Linguistics \\
  Seoul National University \\
  \texttt{snumin44@snu.ac.kr} \\
  \And
  Soyeon Kim \\
  Department of Linguistics \\
  Seoul National University \\
  \texttt{nyong10@snu.ac.kr} \\
  \And
  Chaeyoung Oh \\
  Department of Linguistics \\
  Seoul National University \\
  \texttt{leanne001@snu.ac.kr} \\
  \And
  Jaeyoon Kim \\
  Department of Linguistics \\
  Seoul National University \\
  \texttt{toscour345@snu.ac.kr} \\
  \And
  Hyemi Jo \\
  Department of Linguistics \\
  Seoul National University \\
  \texttt{huimei6361@snu.ac.kr} \\
  \And
  Hyopil Shin \\
  Department of Linguistics \\
  Seoul National University \\
  \texttt{hpshin@snu.ac.kr} \\
}
\renewcommand{\shorttitle}{\textit{arXiv} Template}

\hypersetup{
pdftitle={A template for the arxiv style},
pdfsubject={q-bio.NC, q-bio.QM},
pdfauthor={David S.~Hippocampus, Elias D.~Striatum},
pdfkeywords={First keyword, Second keyword, More},
}

\begin{document}
\maketitle

\begin{abstract}
Pre-trained language models leveraging Transformer architecture have demonstrated remarkable performance across a variety of domains in natural language understanding and generation. In particular, models that are fine-tuned for specific tasks exhibit language representations that are tailored to the nuances of each task. The progression of large language models, particularly those utilizing the Transformer Decoder framework, involves training with an enormous quantity of parameters on expansive datasets. Subsequent to pretraining, these models not only excel in generative tasks but also exhibit exceptional comprehension of natural language. Notably, models undergoing Instruction Tuning after pre-training—which involves adapting to specific templates—have achieved high performance levels, even on tasks that are unseen during initial training. Thus, the development of high-performing Large Language Models (LLMs) increasingly demands the integration of Instruction Tuning into the training process. Within the realm of Korean LLMs, there is a discernible trend toward the public release of models subjected to Instruction Tuning. However, it has been observed that Korean LLMs often rely on datasets either translated from other languages or generated by language models for their training data. Addressing this issue, this paper presents the DaG LLM (David and Goliath Large Language Model), a language model specialized for Korean and fine-tuned through Instruction Tuning across 41 tasks within 13 distinct categories.
\end{abstract}

\keywords{First keyword \and Second keyword \and More}

\section{Introduction}
The introduction of the Transformer architecture by \citet{vaswani2017attention} marked a seminal moment in natural language processing (NLP), setting a new benchmark for subsequent research and advancements. At the heart of the Transformer's innovation are its self-attention mechanisms, which have paved the way for a series of pioneering language models that significantly enhance language understanding and generation capabilities. This surge in progress is exemplified by the GPT series, especially GPT-3, which demonstrated the broad impact of extensive pre-training \citep{vaswani2017attention, brown2020language}. The 2021 debut of InstructGPT spotlighted the adaptability of models fine-tuned for specific tasks \citep{ouyang2022training}. Models like LLaMA \citep{touvron2023llama} and WebGPT \citep{nakano2021webgpt}, designed for internet-derived content, have extended the application scope of language models even further.

Despite these advances, a disparity in resource availability persists among languages, particularly for languages such as Korean. English benefits from an abundance of specialized instruction datasets that facilitate domain-specific model training, while Korean remains underserved. Korean language resources often consist of translations or are generated by large-scale models like those using the ChatGPT API, which may not fully capture the cultural and linguistic subtleties unique to Korean. This highlights an urgent need for native Korean datasets that accurately encompass these nuances to enhance the performance of Korean-targeted language models.

Current Korean language models, such as Eleuther AI's Polyglot-Ko \citep{polyglotko}, Naver Corporation's HyperCLOVA and HyperCLOVAX \citep{hyperclova}, Korea University's KULLM \citep{kullm}, and KoAlpaca, based on Stanford Alpaca \citep{alpaca}, are notable steps forward. However, models developed by major corporations often remain proprietary, while those that are publicly accessible face challenges, most notably their reliance on translated instructional datasets which limit their functionality relative to closed-source counterparts.

This study aims to address these shortcomings through the introduction of the DaG (David and Goliath) project. The project focuses on enhancing the performance of large language models (LLMs) with relatively smaller parameter sets by establishing a systematic process for developing comprehensive Korean instruction datasets over various domains. We introduce the DaG LLM version 1.0, a model adapted from the Polyglot-Ko-5.8b and fine-tuned with a diverse array of instruction datasets covering 41 specific Korean scenarios. The model's optimization process includes efforts to mitigate biases and improve the generation quality inherent to the base model.

Notably, this model differentiates itself by being trained on a diverse range of distinctly Korean datasets, moving away from the typical reliance on translated materials. It seeks to correct dataset biases by ensuring proportional representation and highlights the importance of balanced data in creating robust language models.

This paper offers several significant contributions to the field of NLP, particularly regarding Korean language processing:

\begin{itemize}

    \item \textbf{Development of Korean Instruction Datasets}: We introduce a suite of specifically designed instruction datasets for the Korean language. Spanning 41 tasks in 13 categories, they represent a significant expansion of resources for Korean NLP and address the deficit of instruction-driven datasets for non-English languages.

    \item \textbf{Instruction Tuning for Korean Language Modeling}: The authors present a systematic approach to instruction tuning using the Korean Instruction Datasets. This novel method finely tunes a large language model to enhance its Korean language understanding and generation. This strategy serves as a model for future adaptations in other languages.

    \item \textbf{Balancing and Fair Representation in Training Data}: The paper describes the balancing processing implemented to ensure equitable representation within our training datasets. Such a contribution is essential, as combating biases in AI models is critical for ethical and unbiased language technology development. The DaG LLM ver 1.0 was developed with this in mind, which aids in maintaining fairness and minimizing biases in its outputs.

    \item \textbf{DaG LLM ver 1.0}: We introduce the DaG Large Language Model (LLM) version 1.0, a new model for the Korean language engineered to tackle a wide variety of NLP tasks with enhanced proficiency. Thanks to the detailed instruction datasets used for its training, this model stands as one of the pioneering Korean models subjected to such comprehensive, instruction-driven fine-tuning.

\end{itemize}
\section{Related Work}

\textbf{Advancements in Transformer-based Language Models}

The transformative impact of transformer-based models on the field of natural language processing (NLP) is undeniable, starting with the pioneering work of \citet{vaswani2017attention}, which introduced the foundational Transformer model. This architecture has become the cornerstone for subsequent models that have significantly advanced language understanding. Notably, BERT \citep{devlin2018bert} introduced bidirectional encoder representations from transformers, significantly enhancing the model's ability to capture context within language. As BERT's influence extended across a variety of NLP tasks, it became clear that contextualized embeddings are fundamental to modern language understanding.

Following BERT, the GPT series—initiated by Radford et al.'s GPT \citep{radford2018improving, radford2019language} and culminating with GPT-3 \citep{brown2020language}—further demonstrated the power of transformers, particularly the few-shot learning capabilities of GPT-3 across diverse language tasks. Building upon this, InstructGPT \citep{ouyang2022training} stressed the importance of task-specific fine-tuning and showcased the adaptability of language models.

The evolution of language models has also expanded beyond the textual domain, as seen in LLaMA \citep{touvron2023llama}, which integrates text and imagery, and WebGPT \citep{nakano2021webgpt}, which understands the context of web pages. However, despite these advancements, non-English languages, such as Korean, continue to encounter challenges due to limited resources and support. In this context, contributions such as Polyglot-Ko by Eleuther AI \citep{polyglotko}, HyperCLOVA by Naver \citep{hyperclova}, as well as KoAlpaca and KULLM by Korea University \citep{alpaca, kullm}, represent significant progress.

The contrast in model accessibility, particularly between proprietary models like ChatGPT and publicly available models such as KoAlpaca and KULLM, underscores issues related to resource availability and performance disparities. Addressing these challenges, the current paper introduces the DaG project, which aims to enhance the performance of Large Language Models (LLMs) with fewer parameters, specifically designed for the Korean language. Through extensive fine-tuning on a diverse array of Korean datasets, this project endeavors to mitigate dataset biases while preserving contextual authenticity.

\textbf{Developments in Instruction Tuning}

Instruction tuning has emerged as a critical technique for refining pre-trained language models to comprehend and execute user-provided instructions. Remarkable efforts include aggregating over 60 NLP datasets into instruction format by \citet{instruction}, and developing 52k instructions for fine-tuning the Alpaca model (\citet{alpaca}). Additional advancements have been mirrored in the creation of crowdsourced NLP datasets composed of 61 tasks and 193k instances by researchers \citep{natural_lang_instruction}, and the compilation of 170 English NLP datasets into instruction format by \citet{p3}. The significance of instruction-based datasets has been highlighted by contributions like UnifiedQA (\citet{unifiedqa}), LIMA (\citet{lima}), and instruction sets generated using Chat-GPT \citep{peng}.

Despite these strides, a stark scarcity of such resources exists for non-English languages. This scarcity is especially noticeable in the limited availability of open-source instruction datasets for languages like Korean, highlighted by the Koalpaca dataset's reliance on DeepL for instruction translation. Moreover, the instruction sets for KULLM training frequently utilize translated English datasets, such as those by \citet{peng}, \citet{vicuna}, and \citet{dolly}. An effort to narrow this divide is observed in \citet{kolima}'s work, which involved translating the LIMA dataset into Korean. Nevertheless, original Korean instruction datasets, which do not depend on translations, are scarce and have a narrow focus, underlying the necessity for more comprehensive and accessible instruction datasets in Korean to further the development of Korean Large Language Models.
```

\begin{figure*}[hbt]
    \centering
    \includegraphics[width=1.0\textwidth]{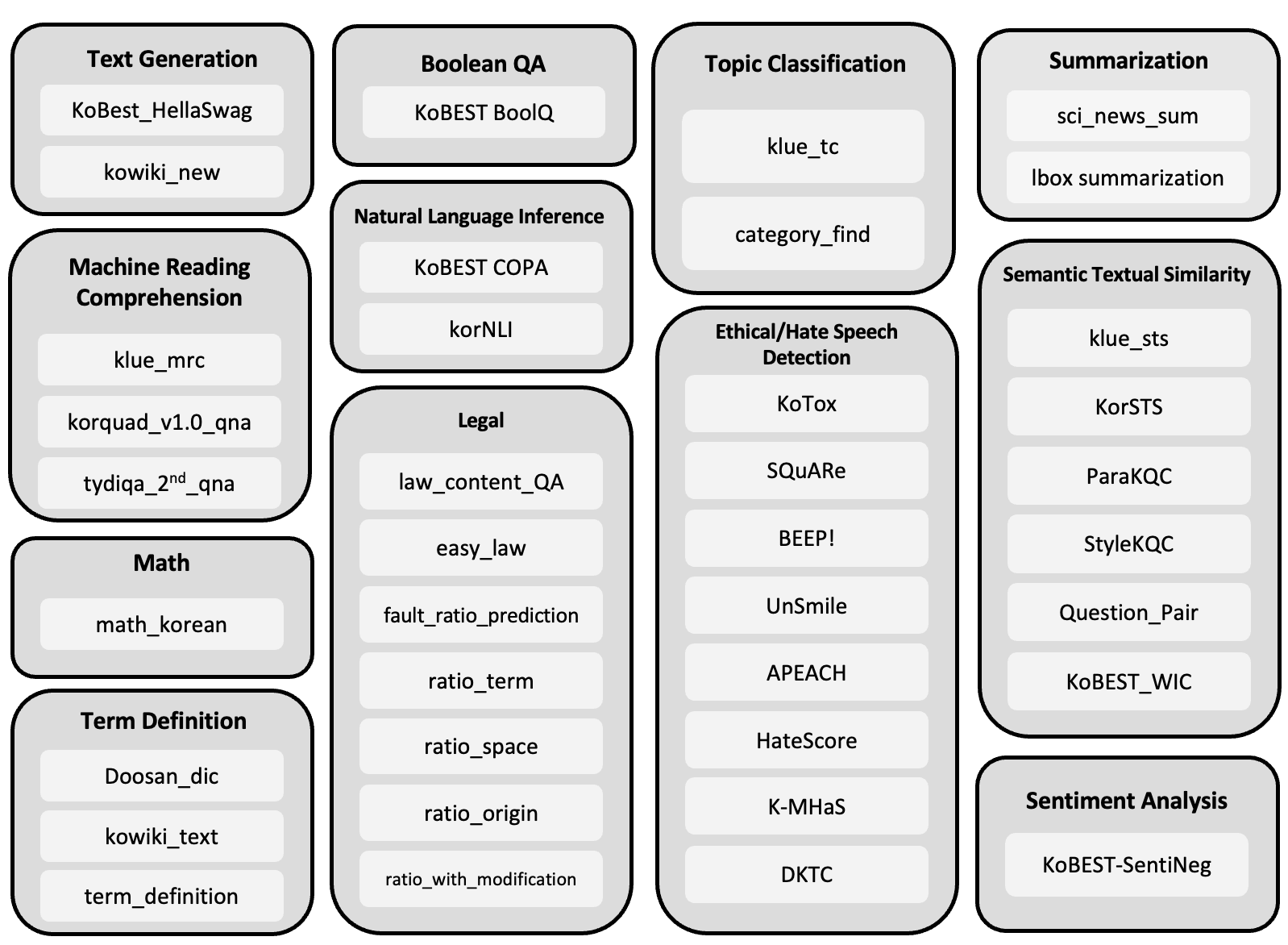} 
    \caption{Model construction and utilization process of DaG LLM ver 1.0}
    \label{fig:datasets}
\end{figure*}

\section{Korean Instruction Datasets}
The meticulous development of Korean Instruction Datasets underpins the efficacy of the DaG LLM v1.0, tailored to execute specific tasks as directed by user-provided instructions. These datasets are derived from tasks spanning 13 essential categories, ranging from straightforward text generation to more complex ethical or legal reasoning. Each category was thoughtfully developed to ensure cultural and linguistic relevance for the Korean-speaking demographic, seeking to address the limitations often faced by non-English LLMs.

The creation of this suite of Korean Instruction Datasets followed a three-step process: selection from open-source corpora, expansion of task categories, and refinement along with template construction. This compilation comprises 41 distinct datasets across 13 categories, each meticulously balanced and crafted not only to meet but also to exceed the capabilities of current instruction-based training methodologies.

\subsection{Categories in the Korean Instruction Datasets}

The diverse categories within the Korean Instruction Datasets offer a rigorous training landscape for DaG LLM v1.0. These categories represent a multifaceted array of tasks that are critical for developing an NLP model that is well-rounded and capable of nuanced understanding and response generation. Herein, we detail the training volume and sub-datasets involved in each category to convey the dataset's richness and its intention for instruction-tuning of models.

\subsubsection{Text Generation}
In text generation, the model tasks include creating contextually relevant and coherent text across various formats, such as narrative construction, story completion, and automated content creation. Instruction-driven methodologies here Foster generative capacity akin to human creativity. The category includes two sub-datasets: KoBEST\_HelleSwag and kowiki\_new, amounting to 13,770 training data points.

\subsubsection{Machine Reading Comprehension}
The machine reading comprehension category includes datasets such as klue\_mrc, korquad\_v1.0\_qna, and tydiqa\_2nd\_qna, with a total of 11,858 data entries. These datasets enhance the model's ability to infer and procure precise information from textual passages to answer questions accurately, reflecting a deep understanding of the text.

\subsubsection{Math}
Comprising math-related tasks, models are expected to interpret and solve mathematical problems articulated in natural language. This testing ground evaluates the model's competency in logical numerical computation as well as linguistic understanding.

\subsubsection{Term Definition}
Term definition datasets challenge the model to provide clear and detailed explanations of specified terms. This skill is fundamental for tasks involving education and knowledge retrieval, necessitating high linguistic clarity.

\subsubsection{Boolean QA}
Boolean QA datasets, represented by KoBEST BoolQ, present binary (true/false) questions requiring models to deduce the correct answers from provided content or world knowledge. This category relies on definitive reasoning capabilities.

\subsubsection{Natural Language Inference}
Natural language inference datasets, like KoBEST COPA and KorNLI, with a combined 10,576 entries, test the model's ability to discern the relational dynamics between sentences, whether entailment, contradiction, or neutrality. These datasets validate the model's capacity for identifying subtle textual implications.

\subsubsection{Legal}
Legal datasets introduce the model to the intricacies of legal discourse, necessitating nuanced comprehension and generation of legalese. With tasks ranging from QA to document analysis, this domain demands exceptional accuracy and contextual awareness.

\subsubsection{Topic Classification}
Topic classification tasks are essential for content organization and retrieval. Guided by datasets such as klue\_tc and category\_find, the model must adeptly assign texts to the correct thematic category, bolstered by 20,000 instances enhancing its training.

\subsubsection{Ethical and Hate Speech Detection}
Ethical and hate speech detection datasets aim to sensitize the model to societal norms, equipping it to identify and mitigate objectionable content. With 56,231 instances spread across datasets like UnSmile, HateScore, and APEACH, the model is honed for responsible and respectful engagement.

\subsubsection{Summarization}
Summarization tasks challenge the model to distill core ideas from extensive texts, requiring comprehensive understanding and synthesis abilities. Instruction-based approaches enable the model to produce succinct summaries, demonstrating effective information processing.

\subsubsection{Semantic Textual Similarity}
Semantic textual similarity, involving datasets like klue\_sts and KorSTS, assesses the degree of semantic correspondence between texts. The model navigates through 28,544 unique examples to master this fundamental aspect of language, beneficial for downstream applications like paraphrase detection and document clustering.

\subsubsection{Sentiment Analysis}
The sentiment analysis category includes datasets such as KoBEST-SentiNeg, where the model identifies and categorizes textual emotional undertones. These datasets prepare the model to detect affective language nuances across 4,446 entries.

The aforementioned categories within the Korean Instruction Datasets provide a robust foundation for instruction-tuning models. By encompassing a diverse ar
ray of NLP tasks, the datasets ensure that models trained on them can competently navigate the complex realm of language with a nuanced, context-aware, and empathetic approach.

\section{Instruction Tuning with Korean Instruction Datasets}

\subsection{Step One: Selection from Open-Source Corpora}
In selecting datasets for instruction tuning, our approach was to leverage open-source materials that offered substantial content for user-centered tasks. Given the generative focus required for the DaG model, our selection favored datasets with clear, actionable guidance and results in the context of the Korean language. We discarded specialized tasks not geared toward general user interaction, streamlining the focus to more universal applications.

\subsection{Step Two: Expanding Task Categories}
In expanding Korean task categories, we deconstructed complex tasks into simpler components, extrapolated new tasks from existing datasets, and broadened the spectrum to cover diverse interactions. For instance, we evolved hate speech detection from a singular dimension to include discrimination and offensive language, each with distinct nuances. Reconfiguration of tasks ensured the instruction dataset was rich and adept at guiding the complex interactions typical in everyday Korean-language usage.

\subsection{Step Three: Refinement and Template Construction}
Our refinement process aimed to construct a dataset fostering nuanced understanding of instructed tasks. We employed \citet{jason}'s template structure to create consistent and varied query iterations. Each task underwent meticulous crafting, ensuring the templates provided differing yet consistent instruction formats. This aimed to acclimate the model to a spectrum of language patterns and structures typical for each task.

During balancing, we placed a cap on dataset entries per task to prevent overfitting—a critical issue for models with fewer parameters. Our curation prioritized balance, leading to a dataset that evenly spans the required task spectrum. The dataset design ensures that DaG LLM v1.0 benefits from a training regime reflecting the variety and richness of tasks it will encounter post-deployment.

The commitment to balance, diversity, and cultural specificity within these datasets underscores the potential of DaG LLM v1.0 to excel where other models may falter, especially regarding the nuanced and often intricate Korean language landscape. It represents not just a technological advancement but a stride toward more inclusive and accessible solutions.
\begin{table}[hbt!]
\centering
\begin{tabular}{|c|c||c|c|}
\specialrule{.1em}{.05em}{.05em} 
Task & Count & Task & Count \\
\specialrule{.1em}{.05em}{.05em} 
klue\_nli\footnote{\url{https://klue-benchmark.com/}} & 5,000 & hatescore & 5,001 \\
sigmorphon\_g2p & 4,500 & klue\_sts & 5,000 \\
dktc & 3,950 & sci\_news\_sum & 50 \\
klue\_mrc & 5,000 & kmhas & 5,000 \\
sae4k\_sum & 5,000 & klue\_tc & 5,000 \\
KoBEST\_BoolQ & 5,000 & pawsx\_paraphr & 5,000 \\
KorNLI\citep{kornli} & 5,000 & KoBEST\_COPA & 4,576 \\
korquad\_v1\_0\_qna & 5,000 & KorSTS\citep{kornli} & 5,000 \\
KoBEST\_HellaSwag & 3,029 & korquad\_v2\_0\_qna & 5,000 \\
Question\_Pair\footnote{\url{https://github.com/songys/Question_pair}} & 5,000 & KoBEST\_SentiNeg & 4,446 \\
tydiqa\_2nd\_qna & 1,858 & StyleKQC\citep{stylekqc}& 5,000 \\ 
KoBEST\_WIC & 5,000 & kor\_nsmc & 10,000 \\
ParaKQC\citep{cho2020discourse} & 5,000 & bias\_comment & 8,367 \\
title\_recommend & 10,000 & kornli & 10,000 \\
hate\_comment & 8,367 & question & 7,576\\
category\_find & 10,000 & korsts & 8,544 \\
nsmc & 10,000 & beep & 5,001 \\
unsmile & 5,001 & kocasm & 5,000 \\
apeach & 3,770 & kowiki\_text & 10,000 \\
kowiki\_new & 10,000 & ratio\_origin & 14,062 \\ 
ratio\_space & 14,062 & ratio\_term & 14,062 \\
\specialrule{.1em}{.05em}{.05em} 
\end{tabular}
\caption{The number of training data per task} 
\label{tab:data_distribution}
\end{table}

\begin{figure*}[t]
    \centering
    \includegraphics[width=1.0\textwidth]{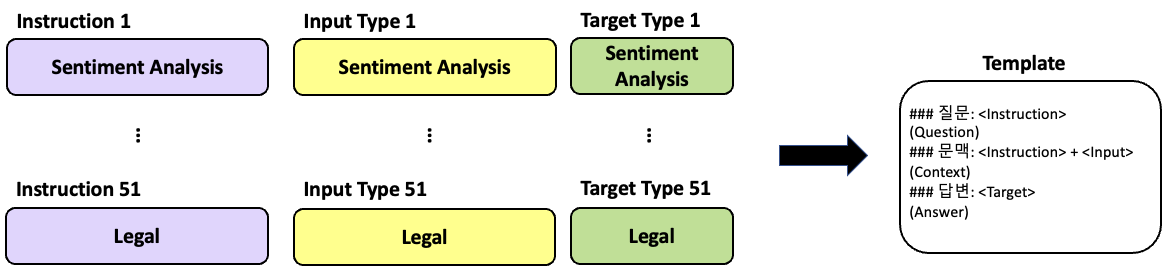} 
    \caption{Model construction and utilization process of DaG LLM ver 1.0}
    \label{fig:fig1}
\end{figure*}

\section{DaG (David and Goliath) LLM ver 1.0}

\subsection{Overview}
The DaG (David and Goliath) LLM ver 1.0 is built upon the Polyglot-Ko-5.8b model, a variant pretrained by EleutherAI that offers a robust foundation for understanding and generating Korean text. Recognized for its efficiency across various NLP tasks even when compared to contemporaneous models of similar size, Polyglot-Ko-5.8b nonetheless reflects a skewed representation of the Korean language - an artifact of its predominantly blog-derived 682.3GB dataset out of a total 863GB training corpus. This inherent bias necessitated an intentional initiative to curate high-quality, balanced, and instruction-centric datasets throughout the pretraining phase to hone in on the desired linguistic nuances and pertinence, as delineated in Chapter 3.

\subsection{Hyper-Parameter Configuration}
The training regimen for DaG LLM is executed by deploying a batch size of 8, bolstered by a Gradient Accumulation setting of 256. This arrangement culminates in an effective batch size of 2048, creating a rich and expansive training ground for the model to iterate across epochs. The learning rate is calibrated to 3e-5, optimizing the model's adaptation to the multifaceted instruction datasets. Leveraging Full Fine-tuning protocols, DaG LLM harnesses the computational might of H-100 GPUs with 80GB of memory each, a testament to the engineering efforts deployed to ensure model robustness and efficiency.

\subsection{Model Utility and Capabilities}
Figure~\ref{fig:fig1} illustrates the DaG LLM ver 1.0's understanding and generation capabilities across a spectrum of NLP tasks, attributed to its training on 41 diverse Korean instruction datasets. It is proficient in both Natural Language Understanding (NLU) and Natural Language Generation (NLG), covering a wide range of activities from text classification and Named Entity Recognition (NER) to document summarization.

DaG LLM's utility extends beyond general-purpose tasks; it is an adept sentiment analyzer, summarization tool, and information retrieval engine. Its versatility is further encompassed by its role as a foundational model for subsequent task-specific enhancements. The model becomes a critical asset for continued learning and specialization, reducing training overheads while enabling a malleable framework for domain-specific tailoring.

\subsection{Instruction Dataset Template}
\begin{table}[hbt!]
\centering
\begin{tabular}{|l|l|}
\hline
\textbf{Type} & \textbf{Content} \\
\hline
Question-based & "\#\#\# Question: <question> \#\#\# Context: <context> <option> \#\#\# Response: <response>" \\
\hline
Instruction-based & "\#\#\# Question: <instruction> \#\#\# Context: <context> <option> \#\#\# Response: <response>" \\
\hline
\end{tabular}
\caption{DaG Instruction Dataset Template}
\label{tab:tab2}
\end{table}

The template is central to the efficacy of instruction tuning in DaG LLM. As reflected in Table~\ref{tab:tab2}, the DaG Instruction Dataset adheres to a dual-structured template: one that prioritizes questions (with optional context inclusion) and another that focuses on the instructional context. Such templates allow the model to diversify its interpretative abilities and enhance its response accuracy during both fine-tuning and Few-Shot Learning regimes. When context is deemed non-essential for the performance of certain tasks, it is accordingly omitted; this flexibility also applies to tasks that do not incorporate explicit questions, ensuring a streamlined approach that aligns with the integral requirements of each instruction.

Overall, DaG LLM ver 1.0 epitomizes a rigorous paradigm within Korean linguistic modeling - endeavoring to set an academic and practical benchmark for executing tailored and broad-ranged language tasks with refined granularity and contextually aware performance. Its inherent training and structural designs promise to foster more accurate, culturally-aligned, and resourceful NLP solutions, fitting the dynamic contours of the Korean language landscape.

\begin{figure*}[hbt]
    \centering
    \includegraphics[width=1.0\textwidth]{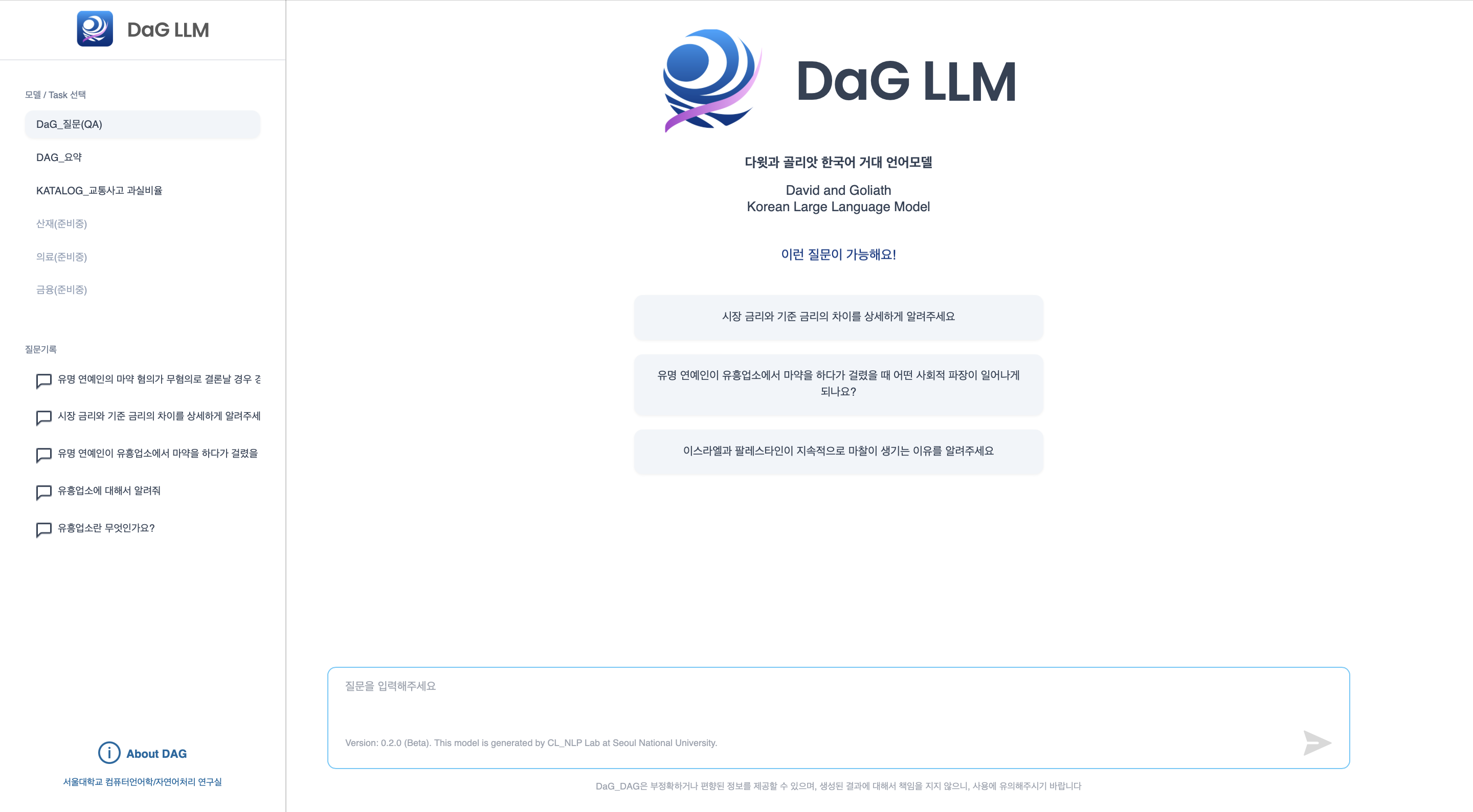} 
    \caption{Web Interface of DaG LLM}
    \label{fig:dag}
\end{figure*}

\subsection{Web Interface Deployment}

To enhance accessibility and practical utility, DaG LLM ver 1.0 has been integrated into a user-friendly web interface available at \url{https://dag.snu.ac.kr}. This web portal is designed to serve as an interactive platform, allowing users to engage with the model in real time and apply its linguistic capabilities across selected domains. The interface provides a streamlined experience, encouraging both academic exploration and everyday use cases.

\subsection{Service Categories}

The current implementation of the DaG LLM interface offers several distinct categories of service, namely Question Answering, Summarization, and KATALOG (Korean Assistant for Traffic Accident Liability Overview Guidance) as seen in \ref{fig:dag}. Each category provides specific functionalities as follows:

\paragraph{Question Answering:} This service leverages the model's comprehension skills to directly answer user queries. It utilizes the model's extensive training on NLU (Natural Language Understanding) tasks to parse input questions and generate accurate and contextually relevant answers. Users can pose questions in a natural conversational manner and receive concise responses from the model, reflecting its sophisticated understanding of diverse topics.

\paragraph{Summarization:} Recognizing the need for succinct information extraction from vast textual content, the Summarization service condenses extensive articles, papers, or reports into shorter versions that retain the original's core message. The model integrates its instruction-tuned summarization training to perform this task with precision, delivering clear and coherent summaries that facilitate quick comprehension of lengthy documents.

\paragraph{KATALOG (Korean Assistant for Traffic Accident Liability Overview Guidance):} The DaG LLM is not only proficient in general-purpose language processing tasks but also exhibits specialized capabilities, particularly through KATALOG. This feature is a prime example of the model's application in specialized domains, requiring intricate legal knowledge and understanding of societal norms.

Here, the model delves into the complexities of traffic accidents within the legal context of Korean jurisdiction. Users can provide detailed accounts of traffic incidents to the model, which then interprets the information to generate a "Final Accident Ratio." This ratio is a critical metric in Korean traffic law, indicating the degree of liability attributed to each party involved in the collision. In addition to providing a numerical liability assessment, the Korean Assistant offers a comprehensive "Accident Analysis," presenting a logical and reasoned dissection of the incident. This analysis forms the backbone of the guidance, elaborating on the circumstances that led to the calculated liability ratio and offering insights that help users correlate the computational output with real-world implications.

Further refinement of the assessment is achieved through the identification of "Modification Factors." These factors take into account external variables that can influence the assignment of fault in traffic accidents, such as environmental conditions, vehicular functionality, and driver behavior at the time of the incident. By considering these, the model provides a nuanced evaluation that mirrors the complex decision-making process of a human legal expert.

To ensure that the guidance is grounded in legal precedent, the model references analogous cases and "Related Judgments." This contextual adherence to established legal outcomes ensures that the model's evaluations are not only logically consistent but also legally sound, taking into account historical legal determinations that resemble the presented scenario.

\begin{figure*}[hbt]
    \centering
    \includegraphics[width=1.0\textwidth]{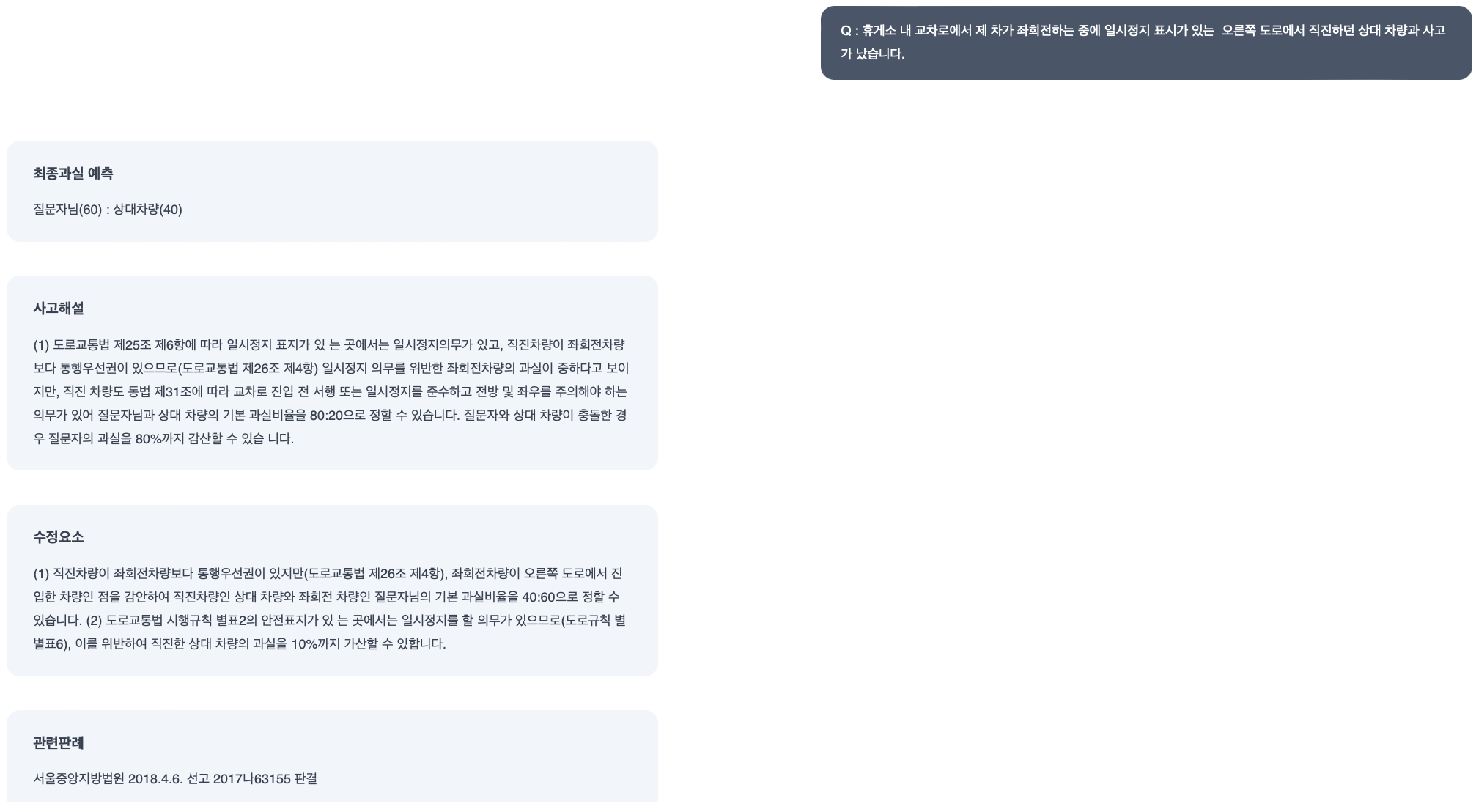} 
    \caption{The Working Example of the KATALOG service}
    \label{fig:katalog}
\end{figure*}

The KATALOG service exemplifies the DaG LLM’s advanced instruction-tuned modeling, demonstrating how language models can execute domain-specific tasks that require high levels of expertise and precision. It underscores the model's flexibility in adapting to specialized knowledge domains and its potential to serve as a valuable tool for individuals navigating complex legal environments (Figure \ref{fig:katalog}).

\subsection{Technical Implementation}

The web interface for DaG LLM is underpinned by a robust backend infrastructure that ensures seamless interaction with the model. Upon user input, the platform communicates with the language model's API, fetching and rendering the generated output in real-time. The user inputs are parsed, and appropriate prompts are constructed based on the defined templates to elicit the desired model responses for each specific service category.

\subsection{Operational Flow}
Interaction with the DaG LLM interface is as follows:

\begin{enumerate}
    \item The user selects a service category and is presented with an input field related to the chosen category.
    \item The user inputs a query, which could be a question, a paragraph for summarization, or details regarding a traffic scenario.
    \item Upon submission, the query is processed by DaG LLM ver 1.0, which employs its tuned instruction datasets and generation capabilities to craft a response.
    \item The response is then delivered to the user through the web interface, providing immediate and accessible insights or answers.
\end{enumerate}

\subsection{Future Directions}

As DaG LLM continues to provide services through the web interface, there are plans to expand the range of available services and enhance the model's precision and scope. User feedback and interaction data will play a crucial role in informing model updates and interface improvements. Moreover, the development team is committed to upholding ethical standards and data privacy, ensuring that user inputs are handled with utmost confidentiality.

Incorporating DaG LLM into a web-based platform represents a significant step toward democratizing sophisticated language processing capabilities, providing users from various domains with a powerful tool to extract, process, and understand complex Korean linguistic data with ease. Through continuous refinement and expansion of services, the DaG LLM ver 1.0 interface aspires to evolve as a cornerstone application for Korean NLP tasks.

\section{Conclusion}

In this paper, we present DaG LLM ver 1.0, an innovative large language model that marks a milestone in the evolving landscape of NLP, especially for the Korean language. Distinguished by its Instruction Tuning with carefully tailored Korean instruction datasets, DaG LLM stands as a paragon of language-specific model training that converges cultural cognizance with advanced linguistic capabilities. With the completion of this model, users benefit from a tool that is precise, versatile, and attuned to the nuances of Korean linguistic phenomena. Its deployment is a leap forward for NLP applications—ranging from nuanced text generation and machine reading comprehension to legal reasoning and ethical speech detection—all within the realm of the Korean language.

\section{Discussion}

This research demonstrates the feasibility and effectiveness of task-specific Instruction Tuning when applied to a large language model with a focus on a non-English language. Through rigorous process design and dataset curation, we affirmed that a well-balanced, culturally informed instructional approach could lead to enhanced model performance in handling diverse Korean language tasks. The DaG LLM ver 1.0 serves not only as a utility model but also as a foundational framework for subsequent research, providing a template for the creation of language models that capture linguistic subtleties and cultural peculiarities inherent in other underrepresented languages.
Furthermore, our work sheds light on the continuous necessity for creating authentic language-specific training datasets, moving away from the traditional reliance on translated resources, propelling the quality of language models towards greater applicability and inclusivity.

\section{Limitations}

While DaG LLM ver 1.0 surmounts several barriers, it is not without constraints. The model's understanding remains influenced by the breadth and depth of the instruction datasets, with its performance potentially hindered by the intrinsic coverage of these datasets. Further challenges include computational resource intensiveness, which could impede swift iteration and expansion, and the potential difficulty in obtaining real-time feedback to fine-tune model responses. Additionally, as cultural nuances continue to evolve, keeping the model's training data up-to-date with contemporary linguistic practices remains an ongoing endeavour.
The scope of the datasets also underscores the intrinsic limitation; a truly comprehensive model would require continuous updates and an ever-expanding dataset that reflects real-world usages and contexts.

\section{Ethical Statement}

This work, although groundbreaking in its aim and execution, is not without ethical considerations. Ensuring that DaG LLM ver 1.0 adheres to respectful and non-discriminatory language usage was of paramount importance throughout the project. We took meticulous care in balancing the datasets to mitigate biases and promote equitable representations.

Nevertheless, there remains the inherent risk that the model may inadvertently generate inappropriate or biased content, a risk prevalent in any language model regardless of the robustness of its training. Thus, we endorse and encourage constant vigilance in monitoring the model's outputs and rectifying any emergent biases. We also advocate for transparency in the model's development and use, upholding the principles of responsible AI.

Admittedly, no model can guarantee complete immunity from ethical dilemmas, especially in complex scenarios that mandate subtle judgment calls. In those instances, human oversight becomes essential. We advise users to employ DaG LLM as a supplemental tool rather than an infallible authority, and we pledge to actively seek out and incorporate feedback to refine and inform the ethical grounding of the model.

In summary, the creation of DaG LLM ver 1.0 represents a significant step towards the creation of nuanced, language-specific language models. While it introduces new possibilities for NLP in the Korean context, it is essential to continue addressing its limitations and ethical dimensions, ensuring that the model remains a beneficial and responsible contributor to the ever-evolving tapestry of AI-driven linguistic technology.

\bibliographystyle{unsrtnat}
\bibliography{references}  






\end{document}